\documentclass[conference]{IEEEtran}
\usepackage[LGR,T1]{fontenc}
\usepackage[latin9]{inputenc}
\usepackage{array}
\usepackage{booktabs}
\usepackage{units}
\usepackage{multirow}
\usepackage{amsmath}
\usepackage{amssymb}
\usepackage{graphicx}
\usepackage{xcolor}

\makeatletter

\providecommand{\tabularnewline}{\\}

\makeatother

\begin{document}
\title{A framework for energy and carbon footprint analysis of distributed
and federated edge learning}
\author{%\author{Stefano Savazzi, Sanaz Kianoush, Vittorio Rampa, Mehdi Bennis}
}
\author{\IEEEauthorblockN{Stefano Savazzi, Sanaz Kianoush, Vittorio Rampa}
\IEEEauthorblockA{Consiglio Nazionale delle Ricerche (CNR)\\
 IEIIT institute, Milano\\
 Email: \{name.surname\}@ieiit.cnr.it} \and \IEEEauthorblockN{Mehdi Bennis} \IEEEauthorblockA{Centre for Wireless Communications\\
 University of Oulu, Finland\\
 Email: mehdi.bennis@oulu.fi}}
\maketitle
\begin{abstract}
Recent advances in distributed learning raise environmental concerns due to the large energy 
needed to train and move data to/from data centers. Novel paradigms, such as 
federated learning (FL), are suitable for decentralized model training across devices or 
silos that simultaneously act as both data producers and learners. Unlike centralized 
learning (CL) techniques, relying on big-data fusion and analytics located in energy hungry 
data centers, in FL scenarios devices collaboratively train their models without sharing 
their private data. This article breaks down and analyzes the main factors that influence 
the environmental footprint of FL policies compared with classical CL/Big-Data algorithms 
running in data centers. The proposed analytical framework takes into account both learning 
and communication energy costs, as well as the carbon equivalent emissions; in addition, it models both vanilla and decentralized FL policies driven by consensus. The framework is evaluated in an industrial setting assuming a real-world robotized workplace. Results show that FL allows remarkable end-to-end energy savings ($30\%\div40\%$) for wireless systems characterized by low bit/Joule efficiency ($50$ kbit/Joule or lower). Consensus-driven FL does not require the parameter server and further reduces emissions in mesh networks ($200$ kbit/Joule). On the other hand, all FL policies are slower to converge when local data are unevenly distributed (often 2x slower than CL). Energy footprint and learning loss can be traded off to optimize efficiency.
\end{abstract}

\IEEEpeerreviewmaketitle{}

\section{Introduction}

Recent advances in machine learning (ML) have revolutionized many 
domains and industrial scenarios. However,
such improvements have been achieved at the cost of large computational
and communication resources, resulting in significant energy
and CO2 (carbon) footprints. Traditional centralized 
learning (CL) requires all training procedures to be conducted inside 
data centers \cite{kone1} that are in charge of collecting training data 
from data producers (\emph{e.g.} sensors, machines and personal devices),
fusing large datasets, and continuously learning from them \cite{commmag}.
Data centers are thus energy-hungry and responsible for significant
carbon emissions that amount to about $15$\% of the global emissions of 
the entire Information and Communication Technology (ICT) ecosystem \cite{carbon}.

An emerging alternative to centralized architectures is federated
learning (FL) \cite{key-7,drl}. Under FL, ML model
parameters, \emph{e.g.} weights and biases $\mathbf{W}$ of Deep
Neural Networks (DNN), are collectively optimized across several resource-constrained edge/fog devices, that act as \emph{both} data producers \emph{and} local
learners. FL distributes the computing task across many devices characterized
by low-power consumption profiles, compared with data centers, and owning small datasets \cite{commmag}.

\begin{figure}[!t]
\centering \includegraphics[scale=0.40]{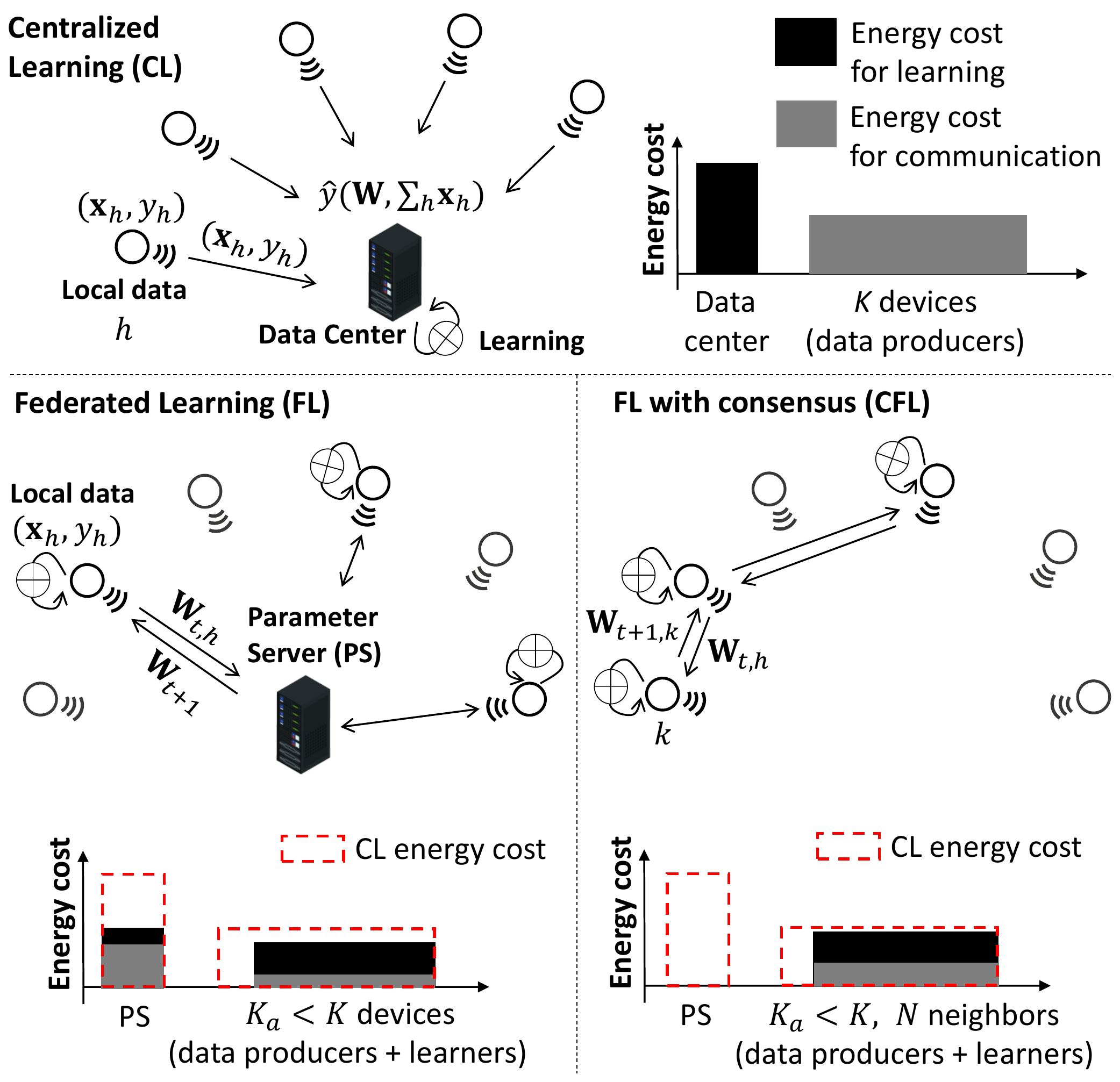} %\par\end{centering}
 \protect\caption{\label{intro} Centralized Learning (CL), Federated Learning (FL)
with Parameter Server (PS), namely federated averaging (FA), and FL
with consensus (\emph{i.e.}, without PS), namely consensus-driven federated
learning (CFL).}
\vspace{-0,4cm}
\end{figure}

As shown in Fig. \ref{intro}, using FL policies, such as
federated averaging \cite{drl}, allows devices to learn a local model under the orchestration of a centralized parameter server (PS). The
PS fuses the received local models to obtain a \emph{global
model} that is fed back to the devices. PS functions are substantially
less energy-hungry compared to CL and can be implemented at the network edge. This suggests that
FL could bring significant reduction in the energy footprints, as the consumption is distributed across devices obviating the need for a large infrastructure for cooling or power delivery. However, vanilla
FL architectures still leverage the server-client architecture which not only
represents a single-point of failure, but also lacks scalability and, if not optimized,
can further increase the energy footprint. To tackle these drawbacks,
recent developments in FL architectures target fully decentralized
solutions relying \emph{solely} on in-network processing, thus replacing
PS functions with a consensus-based federation model. In consensus-based
FL (CFL), the participating devices mutually exchange their local
ML model parameters, possibly via mesh, or device-to-device
(D2D) communication links \cite{commmag}, and implement distributed
weighted averaging \cite{gossip-1,cfa,chen}. Devices might
be either co-located in the same geographic area or distributed.

\textbf{Contributions:} the paper develops a novel framework for the
analysis of energy and carbon footprints in distributed ML, including,
for the first time, comparisons and trade-off considerations about vanilla
FL, consensus-based (CFL) and data center based centralized learning.
Despite an initial attempt to assess the carbon footprint for FL \cite{carbon},
the problem of quantifying an end-to-end analysis of the energy footprint still remains unexplored. To fill this void, we develop an end-to-end framework and validate it using real world data.

The paper is organized as follows: Sections \ref{sec:Energy-footprint-modeling}
and \ref{sec:Carbon-footprint-assessment} describe the framework
for energy consumption and carbon footprint evaluation of different
FL strategies, and the impact of energy efficiency in terms of communication
and computing costs. In Section \ref{sec:A-case-study}, we consider
a case study in a real-world industrial workplace targeting the learning
of a ML model to localize human operators in a human-robot
cooperative manufacturing plant. Carbon emissions are quantified and
discussed in continuous industrial workflow applications
requiring periodic model training updates.

\section{Energy footprint modeling framework \label{sec:Energy-footprint-modeling}}

The proposed framework provides insights into how different components
of the FL architecture, \emph{i.e.} the local learners, the core network
and the PS, contribute to the energy bill. The learning system
consists of $K$ devices and one data center ($k=0$). Each device
$k>0$ has a dataset $\mathcal{E}_{k}$ of (labeled) examples $(\mathbf{x}_{h},y_{h})$
that are typically collected independently. The objective of the learning system is to
train a DNN model $\hat{y}(\mathbf{W};\mathbf{x})$ that transforms
the input data $\mathbf{x}$ into the desired outputs $\hat{y}\in\left\{ y_{c}\right\} _{c=1}^{C}$
where $C$ is the number of the output classes.
Model parameters are specified by the matrix $\mathbf{W}$ \cite{drl}.
The training system uses the examples in $\bigcup_{k=1}^{K}\mathcal{E}_{k}$
to minimize the loss function $\xi(\mathbf{x}_{h},y_{h}|\mathbf{W})$
iteratively, over a pre-defined number $n$ of learning rounds.

Considering a device $k$, the total amount of energy consumed by the learning process can be
broken down into \textrm{computing} and \textrm{communication} components. The energy
cost is thus modelled as a function of the energy $E_{k}^{(\mathrm{C})}$ due to computing
per learning round, and the energy $E_{k,h}^{(\mathrm{T})}$ per correctly
received/transmitted bit over the wireless link ($k,h$).
In particular, the latter can be further broken down into uplink (UL)
communication ($E_{k,0}^{(\mathrm{T})}$) with the data center (or
the PS), and downlink (DL) communication ($E_{0,k}^{(\mathrm{T})}$),
from the PS to the device. The energy cost for communication includes
the power dissipated in the RF front-end, in the conversion, baseband processing 
and transceiver stages. 
We neglect the cost of on-off radio switching. In addition, communication 
energy costs are quantified on average, as routing through the radio access 
and the core network can vary (but might be assumed as stationary apart
from failures or replacements). Finally, the energy $E_{k}^{(\mathrm{C})}$
for computing includes the cost of the learning round, namely the
local gradient-based optimizer and data storage. In what follows,
we quantify the energy cost of model training implemented either inside
the data center (CL) or distributed across multiple devices (FL).
Numerical examples are given in Table \ref{parameters} and in the
case study in Section \ref{sec:A-case-study}.

\subsection{Centralized Learning (CL)}

Under CL, model training is carried out inside the data center $k=0$, while the energy cost per round $E_{0}^{(\mathrm{C})}=P_{0}\cdot T_{0}\cdot B$
depends on the GPU/CPU power consumption $P_{0}$ \cite{carbon},
the time span $T_{0}$ required for processing an individual batch
of data, \emph{i.e.} minimizing the loss $\xi(\cdot|\mathbf{W})$,
and the number $B$ of batches per round. We neglect here the cost of initial dataset loading since it is a one-step process. For $n=n(\overline{\xi})$
rounds, and a target loss $\overline{\xi}$, the total,
end-to-end, energy in Joule {[}J{]} is given by:

\begin{equation}
E_{CL}(\xi)=\gamma\cdot n\cdot E_{0}^{(\mathrm{C})}+\sum_{k=1}^{K}b(\mathcal{E}_{k})\cdot E_{k,0}^{(\mathrm{T})},\label{eq:cl}
\end{equation}
where $\gamma$ is the Power Usage Effectiveness (PUE) of the considered
data center \cite{data_center,cooling}. The cost for UL communication
for data fusion, $\sum_{k=1}^{K}b(\mathcal{E}_{k})\cdot E_{k,0}^{(\mathrm{T})}$,
scales with the data size $b(\mathcal{E}_{k})$ of the $k$-th local
database $\mathcal{E}_{k}$ and the number of devices $K$. PUE $\gamma>1$
accounts for the additional power consumed by the data center infrastructure
for data storage, power delivery and cooling; values are typically
$\gamma=1.1\div1.8$ \cite{cooling}. 
\begin{table}[tp]
\protect\caption{\label{parameters} Computing costs and communication energy efficiency
(EE) values for FL energy/carbon footprint evaluation.}\vspace{-0.2cm}
\begin{centering}
\begin{tabular}{lcc}
\toprule 
\textbf{Parameters} & \textbf{Data center/PS ($k=0$)} & \textbf{Devices ($k\geq1$)}\tabularnewline
\midrule
\midrule 
Comp. $P_{k}$: & $140\,\mathrm{W}$(CPU)$\,+\,42\,\mathrm{W}$(GPU) & $5.1\,\mathrm{W}$ (CPU)\tabularnewline
\midrule 
Batch \negthinspace{}time \negthinspace{}$T_{k}$: & $20$ ms & $190$ ms\tabularnewline
\midrule 
Batches $B$: & $3$ & $3$\tabularnewline
\midrule 
Raw data size: & $K\cdot b(\mathcal{E}_{k})$ MB & $b(\mathcal{E}_{k})\simeq30$ MB\tabularnewline
\midrule 
Model size: & $b(\mathbf{W})=290$ KB & $b(\mathbf{W})=290$ KB\tabularnewline
\midrule 
PUE $\gamma$: & $1.67$ & $1$\tabularnewline
\midrule 
Utilization \negthinspace{}$\beta$: & \multicolumn{2}{l}{$0.1$ (model averaging)}\tabularnewline
\midrule 
ML model: & \multicolumn{2}{l}{DeepMind \cite{deepmind}, $5$ layers, $C=6$. Optimizer: Adam}\tabularnewline
\midrule 
\multirow{4}{*}{Comm. $\mathrm{EE}$:} & \textbf{Downlink (DL):} & \textbf{Uplink (UL):}\tabularnewline
\cmidrule{2-3} \cmidrule{3-3} 
 & $\mathrm{EE_{D}}=0.02\div1$Mb/J & $\mathrm{EE_{U}=0.02\div1}$Mb/J\tabularnewline
\cmidrule{2-3} \cmidrule{3-3} 
 & \multirow{2}{*}{} & \textbf{Mesh or D2D (M):}\tabularnewline
\cmidrule{3-3} 
 &  & $\mathrm{EE_{M}=0.01\div1}$Mb/J\tabularnewline
\midrule 
Comp. $\mathrm{EE}$: & $\mathrm{EE_{C}}=0.9$ round/J & $\mathrm{\tfrac{EE_{C}}{\varphi}}$ round/J, $\varphi=0.22$\tabularnewline
\bottomrule
\end{tabular}
\par\end{centering}
\medskip{}
\vspace{-0.6cm}
\end{table}

\begin{table*}[tp]
	\begin{centering}\protect\caption{\label{carbon_foots} Communication and computing carbon footprints.}\vspace{-0.3cm}
		\begin{tabular}{llll}
			\toprule 
			& \textbf{Communication $C_{\mathrm{C}}$} & \textbf{Computing $C_{\mathrm{L}}$} & \multicolumn{1}{c}{\textbf{Carbon footprint}}\tabularnewline
			\midrule
			\multirow{1}{*}{CL (data center):} & $\sum_{k=1}^{K}b(\mathcal{E}_{k})\cdot \frac{\mathrm{CI}_{k}}{\mathrm{EE_{U}}}$ & $n\cdot\gamma\cdot\frac{\mathrm{CI}_{0}}{\mathrm{EE_{C}}}$ & \multirow{1}{*}{$C_{\mathrm{CL}}=C_{\mathrm{C}}+C_{\mathrm{L}}$}\tabularnewline
			\midrule 
			FL (with PS): $K_{a}\leq K$ & $n\cdot b(\mathbf{W})\cdot\left(\sum_{k=1}^{K_{a}}\frac{\mathrm{CI}_{k}}{\mathrm{EE_{U}}}+\gamma \cdot K \cdot \frac{\mathrm{CI}_{0}}{\mathrm{EE_{D}}}\right)$ & $n\cdot\left(\sum_{k=1}^{K_{a}}\frac{\varphi\cdot \mathrm{CI}_{k}}{\mathrm{EE_{C}}}+\beta\cdot\gamma \cdot \frac{\mathrm{CI}_{0}}{\mathrm{EE_{C}}}\right)$ & $C_{\mathrm{FL}}=C_{\mathrm{C}}+C_{\mathrm{L}}$\tabularnewline
			\midrule 
			CFL : $K_{a}\leq K$, $N\geq1$ & $n\cdot b(\mathbf{W})\cdot\left(\sum_{k=1}^{K_{a}}\frac{N\cdot\mathrm{CI}_{k}}{\mathrm{EE_{M}}}\right)$, & $n\cdot\sum_{k=1}^{K_{a}}\frac{\varphi \cdot \mathrm{CI}_{k}}{\mathrm{EE_{C}}}$ & $C_{\mathrm{CFL}}=C_{\mathrm{C}}+C_{\mathrm{L}}$\tabularnewline
			\bottomrule
		\end{tabular}
		\par\end{centering}
	\medskip{}
	\vspace{-0.6cm}
\end{table*}

\subsection{Federated Learning (FL)}

Unlike CL, FL distributes the learning process across a selected subset $\mathcal{N}_{t}$ 
of $K_{a}<K$ \emph{active} devices as shown in Fig. \ref{intro}. At each round $t$, the local dataset $\mathcal{E}_{k}$ is used
to train a local model $\mathbf{W}_{k,t}$, in order to minimize the
local loss $\xi_{k}$ as $\mathbf{W}_{k,t}=\underset{\mathbf{W}}{\mathrm{arg}\mathrm{min}}\thinspace\xi_{k}(\cdot|\mathbf{W})$.
The local model is then forwarded to the PS \cite{drl} over the UL. The
PS is in charge of updating the global model $\mathbf{W}_{t+1}$ for
the following round $t+1$ through the aggregation of the $K_{a}$
received models \cite{key-7}: $\mathbf{W}_{t+1}=\tfrac{1}{K_{a}}\sum_{k\in\mathcal{N}_{t}}\Gamma_{k}\cdot\mathbf{W}_{k,t}$,
with $\Gamma_{k}=\tfrac{Q_{k}}{Q}$ and ($Q_{k},Q$) being the number
of local and global examples, respectively. The new model $\mathbf{W}_{t+1}$ is finally sent back to the devices over the DL. Other strategies are discussed in \cite{drl}. Notice that, while $K_{a}$ active devices run the local optimizer and share the local model with the
PS on the assigned round, the remaining $K-K_{a}$ devices have their computing hardware turned off, while the communication interface is powered on to decode the updated global model.

For $n$ rounds, now consisting of learning and communication tasks,
the total end-to-end energy includes both devices and PS consumption,
namely:

\begin{equation}
\begin{aligned}E_{FL}(\xi)={} & \gamma\cdot n\cdot\beta\cdot E_{0}^{(\mathrm{C})}\,+\\
{} & +\gamma\cdot\sum_{t=1}^{n}\sum_{k=1}^{K}b(\mathbf{W})\cdot E_{0,k}^{(\mathrm{T})}\,+\\
{} & +\sum_{t=1}^{n}\sum_{k\in\mathcal{N}_{t}}\left[E_{k}^{(\mathrm{C})}+b(\mathbf{W})\cdot E_{k,0}^{(\mathrm{T})}\right]\,.
\end{aligned}
\label{eq:fa}
\end{equation}
PS energy is given by $\beta\cdot E_{0}^{(\mathrm{C})}$ and depends on the
time, $\beta T_{0}$, needed for model averaging. This is considerably
smaller than the batch time $T_{0}$ at the data center (\emph{i.e.}, $\beta \ll 1$). The energy
cost per round for device $k$ is due to the local optimization over the
data batches $\mathcal{E}_{k}$: $E_{k}^{(\mathrm{C})}=P_{k}\cdot B\cdot T_{k}$.
Notice that, while data centers employ high-performance CPUs, GPUs
or other specialized hardware (\emph{e.g.}, NPUs or TPUs), the devices
are usually equipped with embedded low-consumption CPUs or microcontrollers. Thus, it
is reasonable to assume $E_{k}^{(\mathrm{C})}<E_{0}^{(\mathrm{C})}$.
Model size $b(\mathbf{W})$ quantifies the size in bits of model parameters
to be exchanged, which is typically much smaller compared with the raw
data \cite{drl}: $b(\mathbf{W})\ll b(\mathcal{E}_{k})$. In addition,
the parameters size is roughly the same for each device, unless lossy/lossless
compression \cite{digvanalog}\cite{qsgd} is implemented. Sending
data regularly in small batches simplifies medium access control
resource allocation and frame aggregation operations. As shown in
\cite{carbon}, the PUE for all devices is set to $\gamma=1$.

\subsection{Consensus-driven Federated Learning (CFL)}

\label{sec:Decentralized-FL:-gossip}

In decentralized FL driven by consensus, devices mutually exchange the local model parameters
using a low-power distributed mesh network as backbone \cite{commmag,cfa,digvanalog}.
As shown in the example of Fig. \ref{intro}, devices exchange a compressed version \cite{digvanalog,qsgd,GADMM} of their
local models $\mathbf{W}_{k,t}$ following an assigned graph connecting
the learners, and update them by distributed weighted averaging
\cite{cfa,chen}. Let $\mathcal{N}_{k,t}$ be the set that contains
the $N$ chosen neighbors of node $k$ at round $t$, in every new
round ($t>0$) the device updates the local model $\mathbf{W}_{k,t}$
using the parameters $\mathbf{W}_{h,t}$ obtained from the neighbor
device(s) as $\mathbf{W}_{k,t+1}=\mathbf{W}_{k,t}+\sum_{h\in\mathcal{N}_{k,t}}\Gamma_{h}\cdot (\mathbf{W}_{h,t}-\mathbf{W}_{k,t}$).
Weights can be chosen as $\Gamma_{h}=Q_{h}[N \cdot \sum_{h\in\mathcal{N}_{k,t}}Q_{h}]^{-1}$.
Averaging is followed by gradient-based model optimization on $\mathcal{E}_{k}$.

For $K_{a}<K$ active devices in the set $\mathcal{N}_{t}$ and $n$
rounds, the energy footprint is captured only by device consumption:

\begin{equation}
\begin{aligned}E_{CFL}(\xi)={} & \sum_{t=1}^{n}\sum_{k\in\mathcal{N}_{t}}E_{k}^{(\mathrm{C})}+\\
{} & +\sum_{t=1}^{n}\sum_{k\in\mathcal{N}_{t}}\sum_{h\in\mathcal{N}_{k,t}}b(\mathbf{W})\cdot E_{k,h}^{(\mathrm{T})}\,.
\end{aligned}
\label{eq:cfa}
\end{equation}
 The sum $\sum_{h\in\mathcal{N}_{k,t}}b(\mathbf{W})\cdot E_{k,h}^{(\mathrm{T})}$ models the total energy spent by the device $k$ to diffuse the local
model parameters to $N$ selected neighbors at round $t$.

\begin{figure*}[!t]
	\centering \includegraphics[scale=0.24]{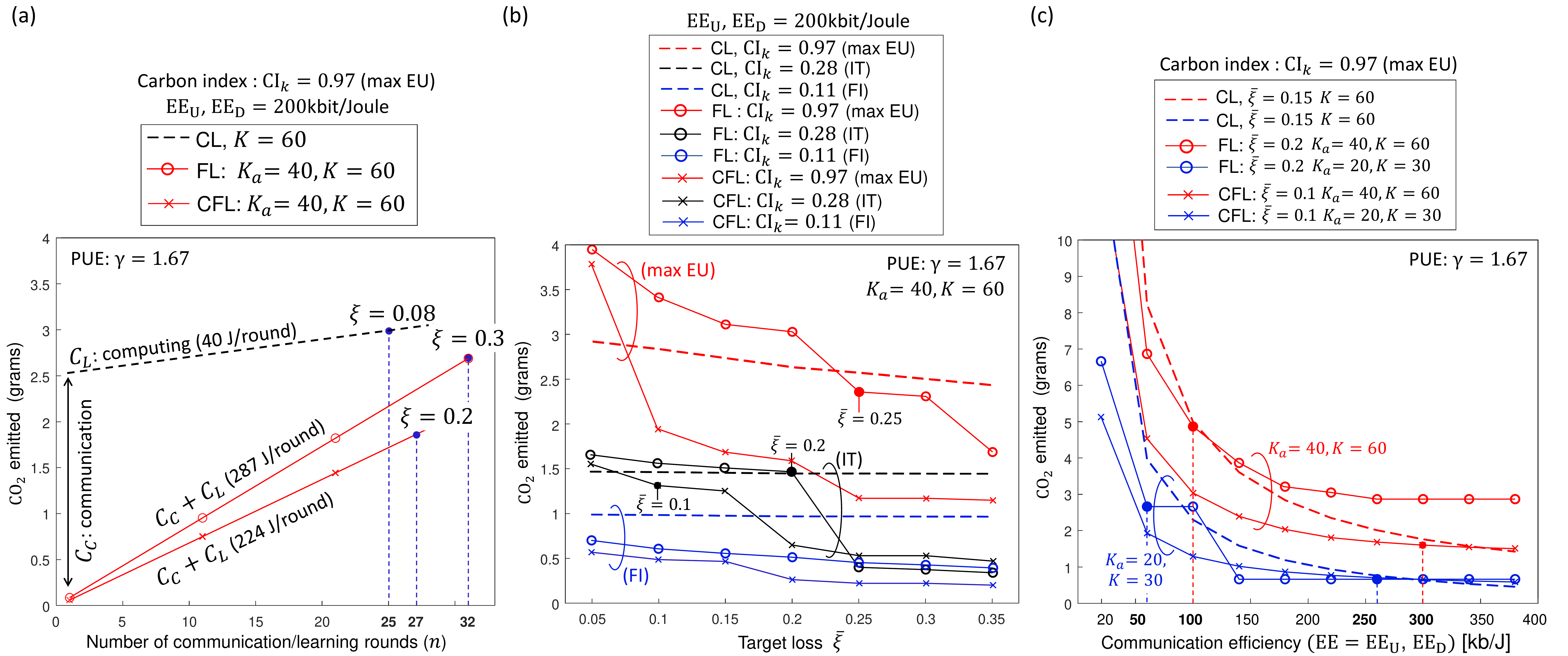} %\par\end{centering}
	\protect\caption{\label{comp1} From left to right. (a) estimated carbon footprints of FL and
		CL for varying number of learning rounds: CL (black) is shown in dashed
		lines for $K=60$ devices, while FL (red with circle markers) and
		CFL (red with cross markers) are shown for $K=60$ devices and $K_{a}=40$
		active ones on each round with $N=1$ neighbors; (b) estimated carbon emissions vs. target loss tradeoff ($K=60$, $K_{a}=40$, $N=1$) and varying CI: max EU (red), Italy (black)
		and Finland (blue); (c) estimated carbon emissions of CL, FL, and CFL for varying
		communication EE ranging from $50$ kbit/J to $400$ kbit/s, and networked devices: $K=30$ $(K_{a}=20)$, and $K=60$ ($K_{a}=40$). Optimal EE below which FL is more carbon efficient than CL is highlighted.}
	\vspace{-0,3cm}
\end{figure*}

\section{Carbon footprint assessment \label{sec:Carbon-footprint-assessment}}

The carbon footprint evaluation assumes that each device $k$, including
the server, is located in a specific geographical region characterized
by a known carbon intensity ($\mathrm{CI}_{k}$) of electricity generation
\cite{key-2}. CI is measured in kg CO2-equivalent emissions per kWh (kgCO2-eq/kWh) 
which quantifies how much carbon emissions are produced
per kilowatt hour of generated electricity. In the following, we consider
the CI figures reported in EU back in 2019 \cite{CI}. Considering the energy
models (\ref{eq:cl})-(\ref{eq:cfa}), carbon
emission is evaluated by multiplying each individual energy contribution,
namely $E_{k}^{(\mathrm{C})}$ and $E_{k,h}^{(\mathrm{T})}$ by the
corresponding intensity values $\mathrm{CI}_{k}$. Carbon footprints
and the proposed framework are summarized in Table \ref{carbon_foots}
for CL ($C_{\mathrm{CL}}$) and FL policies ($C_{\mathrm{FL}}$) and ($C_{\mathrm{CFL}}$).

To analyze the main factors that impact the estimated carbon emissions,
a few simplifications to the energy models (\ref{eq:cl})-(\ref{eq:cfa}) 
are introduced in the following. Communication
and computing costs are quantified on average, in terms of the corresponding
energy efficiencies (EE). Communication EE for DL ($\mathrm{EE}_{\mathrm{D}}=[E_{0,k}^{(\mathrm{T})}]^{-1}$),
UL ($\mathrm{EE}_{\mathrm{U}}=[E_{k,0}^{(\mathrm{T})}]^{-1}$) and
mesh networking ($\mathrm{EE}_{\mathrm{M}}=[E_{k,h}^{(\mathrm{T})}]^{-1}$)
are measured in bit/Joule [bit/J] and describe how much
energy is consumed per correctly received information bit \cite{ee}.
Efficiencies depend on device/server consumption for communication
$P_{\mathrm{T}}$ and net UL/DL or mesh throughput $R$. Depending on network 
implementations, we consider different choices of $\mathrm{EE}_{\mathrm{D}}$, $\mathrm{EE}_{\mathrm{U}}$ and $\mathrm{EE}_{\mathrm{M}}$. The computing efficiency, $\mathrm{EE}_{\mathrm{C}}=[E_{0}^{(\mathrm{C})}]^{-1}$, quantifies the number of rounds per Joule [round/J], namely how much energy per learning round is consumed at the data
center (or PS). Devices equipped with embedded low-consumption CPUs
typically experience a larger time span $T_{k}>T_{0}$ to process
an individual batch of data; on the other hand, they use much lower
power ($P_{k}$). Device computing $\mathrm{EE}$ is typically larger and
modeled here as $\frac{\mathrm{EE_{\mathrm{C}}}}{\varphi}$ with $\varphi=E_{k}^{(\mathrm{C})}/E_{0}^{(\mathrm{C})}<1$. Typical values for communication and computing $\mathrm{EE}$ are in Table \ref{parameters}.

In the proposed FL implementation, the set of $K_{a}$ active FL devices
changes according to a round robin scheduling, other options are proposed in \cite{jointopt}. Considering typical
CFL implementations, such as gossip \cite{gossip-1}, we let the
devices choose up to $N=1$ neighbors per round. When ad-hoc mesh,
or D2D, communication interfaces are not available, the energy cost to implement the generic peer-to-peer link ($k,h$) roughly corresponds to an UL transmission from the source $k$ to
the core network access point (\emph{i.e.}, router), followed by a DL communication from the router(s) to the destination device $h$, namely $E_{k,h}^{(\mathrm{T})}\simeq E_{k,0}^{(\mathrm{T})}+E_{0,h}^{(\mathrm{T})}$,
or equivalently $[\mathrm{EE}_{\mathrm{M}}]^{-1}\simeq [\mathrm{EE}_{\mathrm{D}}]^{-1}+[\mathrm{EE}_{\mathrm{U}}]^{-1}$.
Router can be a host or base-station. In mesh networks, further optimization
via power control \cite{camad} may be also possible depending on
the node deployment. Since devices do not need the router to relay
information to the PS, which may be located in a different country,
substantial energy savings are expected.

\begin{table*}[tp]\protect\caption{\label{simulations} Number of rounds (min-max), communication/computing energy costs and 
		corresponding carbon footprints for selected cases, varying losses $\overline{\xi}$, and IID
		vs. non-IID data distributions. $\mathrm{EE}_{\mathrm{U}}=\mathrm{EE}_{\mathrm{D}}=100$ kbit/J}	 \vspace{-0.3cm}
	\centering \includegraphics[scale=0.415]{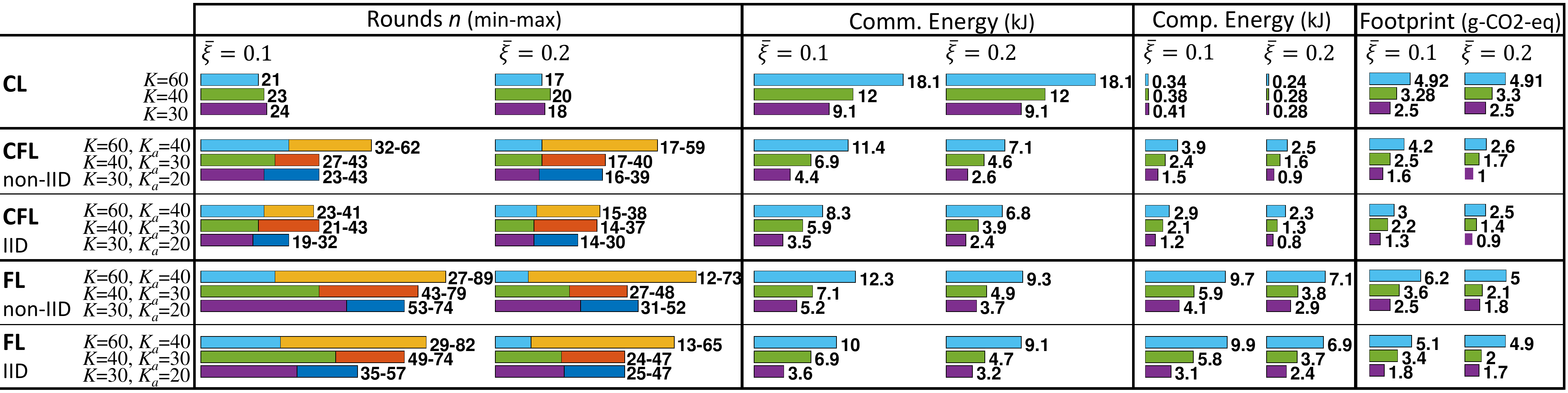} %\par\end{centering}
	
	\medskip{}
	\vspace{-0.6cm}
\end{table*}

\section{Industry 4.0 robotized environment\label{sec:A-case-study}}

According to \cite{trend}, in 2019 industry was responsible for
about $30$\% of the world greenhouse gas emissions. To counter this 
impact, Industry 4.0 (I4.0) and other mitigation policies 
have been recently introduced \cite{mitigation-policies}.
In line with the I4.0 paradigm,
we resort to a common Industrial Internet of Things (IIoT) scenario
where AI-based sensors and machines are interconnected and co-located in the 
same plant \cite{kianoush}. These sensors interact within an industrial workspace
where human workers are co-present. Devices are served by a WiFi (IEEE 802.11ac)
network and a router ($P_{\mathrm{T}}=6$ W \cite{power}) is in charge 
of orchestrating the mesh communication or forwarding to
the data center, or PS.

\subsection{Case study: scenario-dependent setup}

The goal of the training task is to learn a ML model for the detection
(classification) of the position of the human operators sharing the
workspace, namely the human-robot distance $d$ and the direction
of arrival (DOA) $\theta$. Further details about the the robotic
manipulators, the industrial environment and the deployed sensors
are given in \cite{commmag}, \cite{camad}. Input data $\mathbf{x}_{h}$,
available online \cite{dataport}, are range-azimuth maps obtained
from $3$ time-division multiple-input-multiple output (TD-MIMO) frequency
modulated continuous wave (FMCW) radars working in the $77$ GHz band
\cite{kianoush}. During the on-line workflow, position ($d$, $\theta$)
information are obtained from the trained ML model and sent to a programmable
logic controller for robot safety control (\emph{e.g.}, emergency stop
or replanning tasks). The ML model adopted for the classification of the
operator location is a simplified version of the DeepMind \cite{deepmind}. 
It consists of $5$ trainable layers and $3$M parameters, of which $170$k are 
compressed, encoded by $16$ bits and exchanged during FL. Model outputs are 
reduced to $C=6$ for the detection of $6$ subject locations around the robot, 
detailed in \cite{dataport}. Batch times and size of exchanged model parameters $b(\mathbf{W})$ (kB) are reported in Table \ref{parameters}. Adam optimizer is used with a Huber loss \cite{drl}. The number of devices ($K$) is in the range $30\leq K\leq60$, 
data can be identically distributed (IID) or non-IID.
Moreover, $20 \le K_a \le 40$ and $N=1$ are assumed.

Energy and carbon footprints are influenced by data center and device
hardware configurations. The data center hardware consumption is reported
in Table \ref{parameters} and uses CPU (Intel i7 8700K, $3.7$ GHz, $64$ GB)
and GPU (Nvidia Geforce GTX 1060, $1.5$ GHz, $3$ GB). For FL devices, 
we use Raspberry Pi 4 boards based
on a low-power CPU (ARM-Cortex-A72 SoC type BCM2711, $1.5$ GHz, $8$ GB). These
devices can be viewed as a realistic pool of FL learners 
embedded in various IIoT applications. FL is implemented using
Tensorflow v$2.3$ backend (sample code available also in \cite{dataport}). 
In what follows, rather than choosing a specific communication protocol, 
we follow a what-if analysis approach, and thus we quantify the estimated
carbon emissions under the assumption of different DL/UL communication 
efficiencies ($\mathrm{EE}$). Since actual emissions may be larger than 
the estimated ones depending on the specific protocol overhead and 
implementation, we will highlight relative comparisons.

\subsection{Case study: carbon footprint analysis}

Fig. \ref{comp1} provides an estimate of the carbon footprint under
varying settings as detailed in Table \ref{parameters}. Fig. \ref{comp1}(a)
shows the carbon footprint for varying number of learning rounds ($n$), 
comparing CL with $K=60$ devices and FL with $K_{a}=40$. For CL (dashed line), 
an initial energy cost shall be paid for UL raw data transmission, which depends 
on the data size $b(\mathcal{E}_{k})$ and the communication EE; in this example, 
$\mathrm{EE}_{\mathrm{U}}=\mathrm{EE}_{\mathrm{D}}=200$
kbit/J. Next, the energy cost is only due to computing ($40$ J/round),
unless new labelled data are produced by devices before the learning process 
ends on the data center. In contrast to CL, FL footprint depends on communication
\textrm{and} computing energy costs per round. CFL (cross markers) has a cost
of $224$ J/round, smaller than FL, namely $287$ J/round (circle
markers) as PS is not required. Notice that mesh communication is replaced by UL 
and DL WiFi transmissions to/from a router. 

Energy and accuracy loss $\xi$ can be traded off to optimize efficiency. 
For example, CL needs $n=25$ rounds at the data center to achieve a
loss of $\xi=0.08$ and a carbon footprint of $2.9$ gCO2-eq. Model
training should be typically repeated every $3$ hours to track modifications
of the robotic cell layout, which corresponds to a total carbon emission
of $8.4$ equivalent kgCO2-eq per year. CFL trains for more rounds
(here $n=27$) to achieve a slightly larger loss ($\xi=0.2$), but
reduces the emissions down to $1.7$ gCO2-eq, or $4.9$ kgCO2-eq per year,
if training is repeated every $3$ hours. Finally, FL achieves
a similar footprint, however this comes in exchange for a larger validation
loss ($\xi=0.3$) due to communication with the PS. Although not considered here, tuning of model  as well as changing the aggregation strategy 
at the PS \cite{drl} would reduce the training time and thus emissions.

The end-to-end energy cost is investigated in Figs. \ref{comp1}(b) and \ref{comp1}(c).
Energy vs. loss trade-off is first analyzed in Fig. \ref{comp1}(b). We
consider $3$ setups where the data center and the devices are placed
in different geographical areas featuring different carbon indexes
(CIs). In particular, the first scenario (max EU, red) is characterized
by devices located in a region that produces considerable emissions as $\mathrm{CI}_{k}=0.97$ kgCO2-eq/kWh. This corresponds
to the max emission rate in EU \cite{CI}. In the second (IT, black)
and third (FI, blue) scenarios, devices and data center are located
in Italy, $\mathrm{CI}_{k}=0.28$ kgCO2-eq/kWh, and Finland, $\mathrm{CI}_{k}=0.11$
kgCO2-eq/kWh, respectively. When the availability
of green energy is small (\emph{i.e.}, max EU scenario, $\mathrm{CI}_{k}=0.97$),
the learning loss and accuracy must be traded with carbon emissions.
For example, for an amount of gas emission equal, or lower, than CL, the
learning loss of CFL should be increased to $\overline{\xi}=0.1$,
corresponding to an average accuracy of $90\%$. Considering FL, this
should be increased to $\overline{\xi}=0.25$. For smaller carbon
indexes, \emph{i.e.} IT and FI scenarios, the cost per round reduces.
Therefore, FL can train for all the required rounds and experience
the same loss as in CL with considerable emission savings ($30\%\div40\%$
for Finland). A promising roadmap for FL optimization is to let local learners 
contribute to the training process if, or when, green energy, namely small
$\mathrm{CI}_{k}$, is made available.

In Fig. \ref{comp1}(c) we now quantify the carbon emissions of CL,
FL and CFL for varying communication EE, ranging from $\mathrm{EE}_{\mathrm{U}}=\mathrm{EE}_{\mathrm{D}}=50$
kbit/J to $400$ kbit/J, and number of devices, $K=30$ $(K_{a}=20)$, and $K=60$ ($K_{a}=40$). An increase of the network size or a decrease of the network kb/J efficiency cause communication to emit much more CO2 than training. Since FL is more communication efficient as (compressed) model parameters are exchanged, in line with \cite{carbon}, the best operational condition of FL is under limited communication $\mathrm{EE}$ regimes. For the considered scenario, the optimal EE below which FL leaves a smaller carbon footprint than CL is in the range $50\%\div100$ kbit/J for FL ($\overline{\xi}=0.2$) and $250\%\div300$ kbit/J for CFL ($\overline{\xi}=0.1$). Finally, notice that for all cases FL can efficiently operate under $\mathrm{EE}=50$ kbit/J, typically observed in low power communications \cite{tisch}, and 4G/5G NB-IoT \cite{nbiot}. 

Table \ref{simulations} compares the energy and carbon footprints for IID 
and non-IID data distributions. Computing, communication energy costs and 
corresponding carbon emissions for different target losses are evaluated 
with respect to the max EU scenario. Considering FL and CFL, federated computations are now distributed across $K_{a}$ devices, therefore larger computing costs are needed. Non-IID data generally penalizes both FL and CFL as energy consumption
 increases up to $40\%$ in some cases. For example, while CFL with IID data limits 
 the number of required epochs (targeting $\overline{\xi}=0.1$) to a maximum of $n=43$, 
 it is less effective for non-IID distributions as the required rounds now increase 
 up to $n=62$ for some devices. CFL and FL thus experience an increase in energy 
 costs, but CFL still emits lower carbon emissions. More advanced gradient-based 
 CFL methods \cite{cfa} might be considered when data distributions across devices 
 are extremely unbalanced.

\section{Conclusions}

This work developed a framework for the analysis of energy and carbon footprints in distributed and federated learning (FL).
It provides, for the first time, a trade-off analysis between vanilla 
and consensus FL on local datasets, and centralized learning inside the data 
center. A simulation framework has been developed for the performance analysis
over arbitrarily complex wireless network structures. Carbon equivalent 
emissions are quantified and discussed for a continual industrial workflow
monitoring application that tracks the movements of workers inside human-robot 
shared workspaces. The ML model is periodically (re)trained to track changes
in data distributions. In many cases, energy and accuracy should
be traded to optimize FL energy efficiency. Furthermore, by eliminating
the parameter server, as made possible by emerging decentralized FL
architectures, further reducing the energy footprint is a viable solution.
Novel opportunities for energy-aware optimizations are also highlighted. 
These will target the migration of on-device computations where the 
availability of green energy is larger. Finally, FL requires a frequent 
and intensive use of the communication interfaces. This mandates a co-design 
of the federation policy and the communication architecture, rooted in 
the novel 6G paradigms.

\end{document}